\documentclass[letterpaper]{article} % DO NOT CHANGE THIS
\usepackage{aaai24}  % DO NOT CHANGE THIS
\usepackage{times}  % DO NOT CHANGE THIS
\usepackage{helvet}  % DO NOT CHANGE THIS
\usepackage{courier}  % DO NOT CHANGE THIS
\usepackage[hyphens]{url}  % DO NOT CHANGE THIS
\usepackage{graphicx} % DO NOT CHANGE THIS
\usepackage{natbib}  % DO NOT CHANGE THIS AND DO NOT ADD ANY OPTIONS TO IT
\usepackage{caption} % DO NOT CHANGE THIS AND DO NOT ADD ANY OPTIONS TO IT
\usepackage{algorithm}
\usepackage{algorithmic}

\usepackage{newfloat}
\usepackage{listings}
\DeclareCaptionStyle{ruled}{labelfont=normalfont,labelsep=colon,strut=off} % DO NOT CHANGE THIS
\floatstyle{ruled}
\newfloat{listing}{tb}{lst}{}
\floatname{listing}{Listing}
\usepackage{hyperref} % -- This package is specifically forbidden
\hypersetup{
    colorlinks=true,
    linkcolor=blue,
    citecolor=blue,
    filecolor=magenta,
    urlcolor=blue
    }

\usepackage{amsmath, amsfonts} 
\usepackage{cuted}
\usepackage{xspace}
\usepackage{booktabs}
\usepackage{multirow}
\usepackage{array}
\newcolumntype{M}[1]{>{\centering\arraybackslash}p{#1}}

\newcommand{\quotes}[1]{``#1''}
\newcommand{\ua}[1]{\tiny{($\uparrow$#1)}}
\newcommand{\da}[1]{\tiny{($\downarrow$#1)}}

\title{MotionMix: Weakly-Supervised Diffusion for Controllable Motion Generation}
\author{
    Nhat M. Hoang\textsuperscript{\rm 1,2}\thanks{Work done during an internship at Huawei}, Kehong Gong\textsuperscript{\rm 1}\thanks{Corresponding author}, Chuan Guo\textsuperscript{\rm 1}\footnotemark[1], Michael Bi Mi\textsuperscript{\rm 1}
}
\affiliations{
    \textsuperscript{\rm 1}Huawei Technologies Co., Ltd.,\\
    \textsuperscript{\rm 2}Nanyang Technological University \\
    nhat005@e.ntu.edu.sg, gongkehong@u.nus.edu, cguo2@ualberta.ca, michaelbimi@yahoo.com
}

\usepackage{bibentry}
\begin{document}

\maketitle

\begin{strip}
\centering
\captionsetup{skip=0.5\baselineskip}
\includegraphics[width=0.85\textwidth]{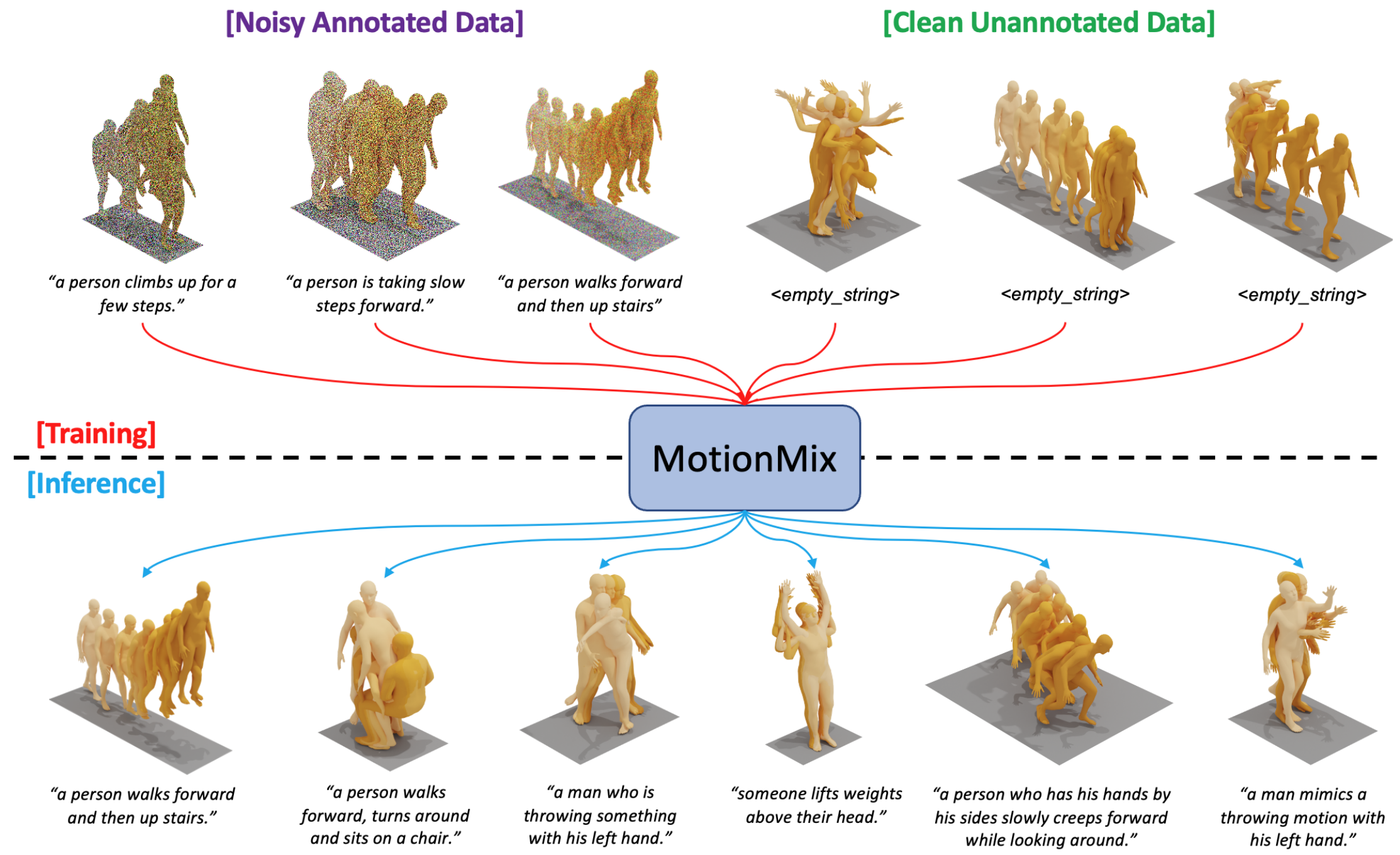}
    \captionof{figure}{Examples of applying MotionMix on text-to-motion generation. Unlike previous works, our training data are only comprised of \textit{noisy annotated motions} and \textit{unannotated motions}. \url{https://nhathoang2002.github.io/MotionMix-page/}}
\label{fig:intro-fig}
\end{strip}

\begin{abstract}
Controllable generation of 3D human motions becomes an important topic as the world embraces digital transformation. Existing works, though making promising progress with the advent of diffusion models, heavily rely on meticulously captured and annotated (e.g., text) high-quality motion corpus, a resource-intensive endeavor in the real world. This motivates our proposed \textbf{MotionMix}, a simple yet effective weakly-supervised diffusion model that leverages both \textit{noisy} and \textit{unannotated} motion sequences. Specifically, we separate the denoising objectives of a diffusion model into two stages: obtaining conditional rough motion approximations in the initial $T-T^*$ steps by learning the noisy annotated motions, followed by the unconditional refinement of these preliminary motions during the last $T^*$ steps using unannotated motions. Notably, though learning from two sources of imperfect data, our model does not compromise motion generation quality compared to fully supervised approaches that access gold data. Extensive experiments on several benchmarks demonstrate that our MotionMix, as a versatile framework, consistently achieves state-of-the-art performances on text-to-motion, action-to-motion, and music-to-dance tasks.
\end{abstract}

\section{Introduction}
\label{sec:intro}

The rapidly arising attention and interest in digital humans bring up the great demand for human motion generation, in a wide range of fields such as industrial game and movie animation \cite{Ling2020CharacterCU}, human-machine interaction \cite{Koppula2013AnticipatingHA}, VR/AR and metaverse development \cite{Lee2021AllON}. Over the years, automated generation of human motions that align with user preferences, spanning aspects such as prefix poses \cite{HernandezRuiz2018HumanMP, Guo2022BackTM}, action classes \cite{actor, Cervantes2022ImplicitNR}, textual descriptions \cite{temos, Language2Pose, mdm}, or music \cite{Aristidou2021RhythmIA, bailando, tm2d}, has been a focal point of research. Recently, building upon the advancement of diffusion models, human motion generation has experienced a notable improvement in quality and controllability. However, these prior diffusion models are commonly trained on well-crafted motions that come with explicit annotations like textual descriptions. While capturing motions from the real world is a laborious effort, annotating these motion sequences further urges the matter. 

In contrast, motions with lower fidelity or fewer annotations are more accessible in the real world. For example, 3D human motions are readily extracted from monocular videos through video-based pose estimation \cite{Kanazawa2017EndtoEndRO, Kocabas2019VIBEVI, Choutas2020MonocularEB}. Meanwhile, a wealth of unannotated motion sequences, such as those available from Mixamo \cite{mixamo} and AMASS \cite{amass}, remains largely untapped. This brings up the question we are investigating in this work, as illustrated in Figure~\ref{fig:intro-fig}. Can we learn reliable diffusion models for controllable motion generation based on the supervision of \textit{noisy} and the \textit{unannotated} motion sequences?

\begin{figure*}[t!]
\centering
    \includegraphics[width=0.99\textwidth]{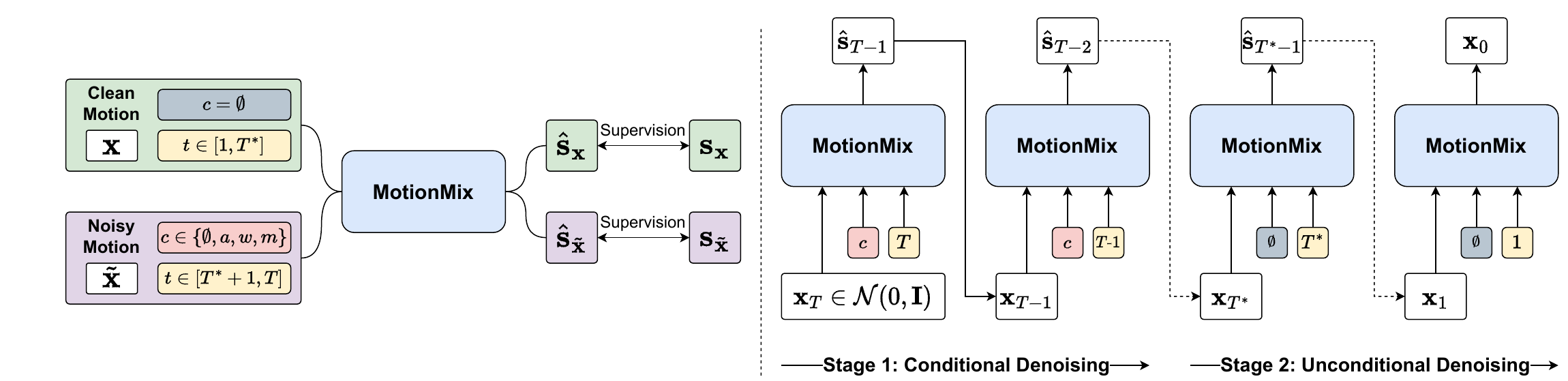}
    \caption{
        (Left) Training Process. The model is trained with a mixture of noisy and clean data. A noise timestep in ranges of $[1, T^*]$ and $[T^*+1, T]$ is sampled respectively for each clean and noisy data. Here, $T^*$ is a denoising pivot that determines the starting point from which the diffusion model refines the noisy motion sequences into clean ones without any guidance. (Right) Sampling Process. The sampling process consists of two stages. In Stage-1 from timestep $T$ to $T^*+1$, the model generates the rough motion approximations, guided by the conditional input $c$. In Stage-2 from timestep $T^*$ to $1$, the model refines these approximations to high-quality motion sequences while the input $c$ is masked.
}
\label{fig:overview-pipeline}
\end{figure*}

Fortunately, with the inherent denoising mechanism of diffusion models, we are able to answer this question with a simple yet effective solution that applies separate diffusion steps regarding the source of training motion data, referred to as \textbf{MotionMix}. To demonstrate our application and approach, we split each gold annotated motion dataset into two halves: the first half of the motions are injected with random-scale Gaussian noises (noisy half), and the second half is deprived of annotations (clean half). As in Figure~\ref{fig:overview-pipeline}, the diffusion model bases on the clean samples for diffusion steps in $[1, T^*]$, with condition input erased. Meanwhile, noisy motions supervise the model with explicit conditions for the rest of steps $[T^*+1, T]$. Note $T^*$ is an experimental hyper-parameter, with its role analyzed in later ablation studies. Our key insight is that, during sampling, starting from Gaussian noises, the model first produces rough motion approximations with conditional guidance in the initial $T-T^*$ steps; afterward, these rough approximations are further refined by unconditional sampling in the last $T^*$ steps. Yet learning with weak supervision signals, our proposed MotionMix empirically facilitates motion generation with higher quality than fully supervised models on multiple applications.  Benefiting from the conciseness of design, MotionMix finds its place in many applications. In this work, we thoroughly examine the effectiveness and flexibility of the proposed approach through extensive experiments on benchmarks of text-to-motion, music-to-dance, and action-to-motion tasks. 

The main contributions of our work can be summarized as follows:

$\bullet$ We present \textbf{MotionMix}, the first weakly-supervised approach for conditional diffusion models that utilizes both \textit{noisy annotated} and \textit{clean unannotated} motion sequences simultaneously.

$\bullet$ We demonstrate that by training with these two sources of data simultaneously, \textbf{MotionMix} can improve upon prior state-of-the-art motion diffusion models across various tasks and benchmarks, without any conflict.

$\bullet$ Our approach opens new avenues for addressing the scarcity of clean and annotated motion sequences, paving the way for scaling up future research by effectively harnessing available motion resources.

\section{Related Work}
\label{sec:related-work}

\subsection{Weakly-Supervised Learning}
\label{ssec:related-weakly-supervised}
To tackle the limited availability of annotated data, researchers have been exploring the use of semi-supervised generative models, using both annotated and unannotated data \cite{Kingma2014SemisupervisedLW, Li2017TripleGA, Lucic2019HighFidelityIG}. However, the investigation of semi-supervised diffusion models remains limited \cite{You2023DiffusionMA}, possibly due to the significant performance gap observed between conditional and unconditional diffusion models \cite{Bao2022WhyAC, Dhariwal2021DiffusionMB, mdm}. Moreover, many state-of-the-art models, such as Stable Diffusion \cite{Rombach2021HighResolutionIS}, implicitly assume the availability of abundant annotated data for training \cite{Chang2023OnTD, Kawar2023GSUREBasedDM}. This assumption poses a challenge when acquiring high-quality annotated data is expensive, particularly in the case of 3D human motion data.

Recent interest has emerged in developing data-efficient approaches for training conditional diffusion models with low-quality data \cite{Daras2023AmbientDL, Kawar2023GSUREBasedDM}, or utilizing unsupervised \cite{Tur2023ExploringDM}, semi-supervised \cite{You2023DiffusionMA}, self-supervised methods \cite{Miao2023DDS2MSD}. These approaches have exhibited promising results across various domains and hold potential for future exploration of diffusion models when handling limited annotated data. However, in the domain of human motion generation, efforts toward these approaches have been even more limited. One related work, Make-An-Animation \cite{maa}, trains a diffusion model utilizing unannotated motions in a semi-supervised setting. In contrast, our work introduces a unique aspect by training with noisy annotated motion and clean unannotated motion.

\subsection{Conditional Motion Generation}
\label{ssec:related-motion-generation}
Over the years, human motion generation has been extensively studied using various signals, including prefix poses \cite{HernandezRuiz2018HumanMP, Guo2022BackTM, actor}, action classes \cite{humanact, actor, Cervantes2022ImplicitNR}, textual descriptions \cite{humanml3d, temos, Ghosh2021SynthesisOC, t2mt, Language2Pose, Text2Gestures}, or music \cite{Li2020LearningTG, Aristidou2021RhythmIA, aistpp, bailando, tm2d}. However, it is non-trivial for these methods to align the distributions of motion sequences and conditions such as natural languages or speech \cite{mld}. Diffusion models resolve this problem using a dedicated multi-step gradual diffuse and denosing process\cite{Ramesh2022HierarchicalTI, Saharia2022PhotorealisticTD, Ho2022ImagenVH}. Recent advancements, such as MDM \cite{mdm}, MotionDiffuse \cite{motiondiffuse}, MLD \cite{mld}, have demonstrated the ability of diffusion-based models to generate plausible human motion, guided by textual descriptions or action classes. In the domain of music, EDGE \cite{edge} showcased high-quality dance generation in diverse music categories. Nevertheless, these works still rely on high-quality motion datasets with annotated guidance.

\section{Method}
\label{sec:method}

\subsection{Problem Formulation}
\label{ssec:problem-formulation}
Conditional motion generation involves generating high-quality and diverse human motion sequences based on a desired conditional input $c$. This input can take various forms, such as a textual description $w^{1:N}$ of $N$ words \cite{humanml3d}, an action class $a \in A$ \cite{humanact}, music audio $m$ \cite{aistpp}, or even an empty condition $c = \emptyset$ (unconditional input) \cite{modi}. Our goal is to train a diffusion model in a weakly-supervised manner, using both noisy motion sequences with conditional inputs $c = \{\emptyset, a, w, c\}$ (where $\emptyset$ is used when the classifier-free guidance \cite{ho2022classifierfree} is applied) and clean motion sequences with unconditional input $c = \emptyset$. Despite being trained with noisy motions, our model can consistently generate plausible motion sequences. To achieve this, we propose a two-stage reverse process, as illustrated in Figure~\ref{fig:overview-pipeline}.

\subsection{Diffusion Probabilistic Model}
\label{ssec:diffusion-model}
The general idea of a diffusion model, as defined by the denoising diffusion probabilistic model (DDPM) \cite{ddpm}, is to design a \textit{diffusion process} that gradually adds noise to a data sample and trains a neural model to learn a \textit{reverse process} of denoising it back to a clean sample. Specifically, the diffusion process can be modeled as a Markov noising process with $\{\mathbf{x}_t \}_{t=0}^T$ where $\mathbf{x}_0 \sim p(x)$ is the clean sample drawn from the data distribution. The noised $\mathbf{x}_t$ is obtained by applying Gaussian noise $\boldsymbol{\epsilon}_t$ to $\mathbf{x}_0$ through the posterior:

\begin{equation}
  q(\mathbf{x}_t  | \mathbf{x}_0) = \mathcal{N}(\mathbf{x}_t; \sqrt{\bar{\alpha}_t} \mathbf{x}_0, (1 - \bar{\alpha}_t)\mathbf{I})
  \label{eq:forward-process}
\end{equation}

\noindent where $\bar{\alpha}_t \in (0, 1)$ are constants which follow a monotonically decreasing scheduler. Thus, when $\bar{\alpha}_t$ is small enough, we can approximate $\mathbf{x}_T \sim \mathcal{N}(0, \mathbf{I})$.

In the reverse process, given the condition $c$, a neural model $f_\theta$ is trained to estimate the clean sample $\mathbf{x}_0$ \cite{ramesh2022hierarchical} or the added noise $\epsilon_t$ \cite{ddpm} for all $t$. The model parameters $\theta$ are optimized using the \quotes{simple} objective introduced by \citeauthor{ddpm}:

\begin{equation}
  \mathcal{L}_\text{simple} = \mathbb{E}_{t \sim [1, T], \mathbf{s}_t} \Big[\|\mathbf{s}_t - f_\theta(\mathbf{x}_t, t, c)\|^2 \Big]
  \label{eq:simple-objective}
\end{equation}

\noindent  where the target objective $\mathbf{s}_t$ refers to either $\mathbf{x}_0$ or $\boldsymbol{\epsilon}_t$ for ease of notation.

\subsection{Training}
\label{ssec:training}
We propose a novel weakly-supervised learning approach that enables a diffusion model to effectively utilize both noisy and clean motion sequences. During the training phase, we construct batches comprising both noisy and clean samples, each coupled with a corresponding guidance condition $c$, as further detailed in Subsection~\ref{ssec:data-preparation}. To learn the denoising process, we apply the diffusion process to this batch using Equation~\ref{eq:forward-process} with varying noise timesteps. In contrary to the conventional training, where both noisy and clean motion sequences are treated as the ground truth $x_0$ with diffusion steps spanning $[1, T]$, our approach adopts separate ranges for different data types. For noisy samples, we randomly select noise timesteps $t \in [T^* + 1, T]$, while for clean samples, we confine them to $t \in [1, T^*]$. Here, $T^*$ serves as a denoising pivot, determining when the diffusion model starts refining noisy motion sequences into cleaner versions. This pivot is especially crucial in real-world applications, where motion capture data might be corrupted by noise due to diverse factors. This denoising strategy for \textit{noisy motions} draws inspiration from \cite{nie2022diffusion}, which purified adversarial images by diffusing them up to a specific timestep $T^*$ before denoising to clean images. The determination of $T^*$ typically relies on empirical estimation, its impact on generation quality is further analyzed in Table~\ref{tab:ablation-denoising-pivot}.

Through this training process, the model becomes adept at generating initial rough motions from $T$ to $T^* + 1$, and subsequently refining these rough motions into high-quality ones from $T^*$ to $1$. By dividing into two distinct time ranges, the model can effectively learn from both noisy and clean motion sequences as ground truth without any conflict.

\subsection{Two-stage Sampling and Guidance}
\label{ssec:sampling}
Our approach introduces a modification to the conventional DDPM sampling procedure, which commonly relies on the same explicit conditional input $c$ to guide the denoising operation at each time step $t$, initiating from $T$ and denoising back to the subsequent time step $t-1$ until reaching $t=0$. However, it is important to note that our work specifically focuses on clean, unannotated samples. As discussed in Subsection~\ref{ssec:training}, these samples are trained using an identical guidance condition $c = \emptyset$ confined within the time interval $[1, T^*]$. Consequently, if the conventional DDPM sampling process is employed within this temporal range, it could potentially lead to jittering or the generation of unrealistic motions. This occurs because the model is not trained to handle varying conditions within this specific range. To tackle this issue, we adopt a distinct strategy to align the sampling process accordingly. Specifically, when the model reaches the denoising pivot $T^*$ during the sampling, we substitute the conditional input with $c = \emptyset$ starting from $T^*$.

In the case of using classifier-free guidance \cite{ho2022classifierfree}, guided inference is employed for all $t$, which involves generating motion samples through a weighted sum of unconditionally and conditionally generated samples:

\begin{equation}
  \hat{\mathbf{s}}(\mathbf{x}_t, t, c) = w \cdot f_\theta(\mathbf{x}_t, t, c) + (1-w) \cdot f_\theta(\mathbf{x}_t, t, \emptyset)
  \label{eq:guided-inference}
\end{equation}

\noindent where $w$ is the guidance weight during sampling.

\subsection{Data Preparation}
\label{ssec:data-preparation}
To facilitate our setting, we randomly partition an existing training dataset into two subsets. In one subset, we retain the annotated condition and introduce noise to the motion sequences to approximate the real noisy samples. In the other subset, we reserve the cleanliness of the data and discard the annotated conditions by replacing them as $c = \emptyset$.

Motivated by the use of Gaussian noises in approximating noisy samples in previous works \cite{PoseNDF, MotionDVAE}, we apply the Equation~\ref{eq:forward-process} to gradually introduce noise to the clean samples. Since the precise noise schedule in real-world motion capture data is unknown, we address this uncertainty by applying a random noising step sampled from the range $[T_1, T_2]$, where $T_1$ and $T_2$ are hyperparameters simulating the level of disruption in real noisy motions. Interestingly, our experiments (Tab.~\ref{tab:ablation-noisy-range}) show that neither smaller value of $T_1$, $T_2$ nor small $T_2-T_1$ relates to better final performance. Due to page limit, examples of noisy motions for training are delegated to present in supplementary videos.

It is worth noting that the processes of dividing the training dataset and preparing noisy samples, and unannotated samples only take place on the side of the training dataset. The remaining evaluation dataset, diffusion models, and training process are kept unchanged as in previous works.

\section{Experiments}
\label{sec:experiments}
We thoroughly experiment our MotionMix in diverse tasks using different conditional motion generation diffusion models as backbones: \textbf{(1)} MDM \cite{mdm} for text-to-motion task on HumanML3D \cite{humanml3d}, KIT-ML \cite{kit}, as well as action-to-motion task on HumanAct12 \cite{humanact} and UESTC \cite{uestc}; \textbf{(2)} MotionDiffuse \cite{motiondiffuse} for text-to-motion task; and \textbf{(3)} EDGE \cite{edge} for music-to-dance task on AIST++ \cite{aistpp}. 

\subsection{Models}
\label{ssec:models}
\noindent \textbf{$\bullet$ MDM \cite{mdm}}
MDM is a lightweight diffusion model that utilizes a transformer encoder-only architecture \cite{Vaswani2017AttentionIA}. Its training objective is to estimate the clean sample $\mathbf{x}_0$ \cite{ramesh2022hierarchical}. In the text-to-motion task, MDM encodes the text description $c = w^{1:N}$ using a frozen CLIP-VIT-B/32. During training, classifier-free guidance \cite{ho2022classifierfree} is employed by randomly masking the condition with $c = \emptyset$ with a probability of $10\%$. Meanwhile, in the action-to-motion task, the conditioning $c = a$ is projected to a linear action embedding, and the classifier-free guidance is not applied. Additionally, three geometric losses are incorporated as training constraints for this task.

\noindent \textbf{$\bullet$ MotionDiffuse \cite{motiondiffuse}}
MotionDiffuse employs a series of transformer decoder layers \cite{Vaswani2017AttentionIA} and incorporates a frozen CLIP-VIT-B/32 for text description encoding. However, in contrast to MDM, MotionDiffuse focuses on estimating the noise $\boldsymbol{\epsilon}$ as its training objective and does not incorporate the classifier-free guidance \cite{ho2022classifierfree}.

\noindent \textbf{$\bullet$ EDGE \cite{edge}.}
EDGE shares similarities with MDM in terms of its transformer encoder-only architecture \cite{Vaswani2017AttentionIA} and the adoption of geometric losses for the music-to-dance task. In addition, the authors introduced a novel Contact Consistency Loss to enhance foot contact prediction control. In the case of music conditioning, EDGE utilizes a pre-trained Jukebox model \cite{jukebox} to extract audio features $m$ from music, which then serve as conditioning input $c = m$. During inference, the approach incorporates classifier-free guidance \cite{ho2022classifierfree} with a masking probability of $25\%$.

\subsection{Text-to-motion}
\label{ssec:t2m}

\noindent \textbf{$\bullet$ Datasets.} Two leading benchmarks used for text-driven motion generation are HumanML3D \cite{humanml3d} and KIT-ML \cite{kit}. The KIT-ML dataset provides 6,353 textual descriptions corresponding to 3,911 motion sequences, while the HumanML3D dataset combines 14,616 motion sequences from HumanAct12 \cite{humanact} and AMASS \cite{amass}, along with 44,970 sequence-level textual descriptions. As suggested by \citeauthor{humanml3d}, we adopt a redundant motion representation that concatenates root velocities, root height, local joint positions, velocities, rotations, and the binary labels of foot contact. This representation, denoted as $\mathbf{x} \in \mathbb{R}^{N\times D}$, is used for both HumanML3D and KIT-ML, with $D$ being the dimension of the pose vector and is equal to 263 for HumanML3D or 251 for KIT-ML. This motion representation is also employed in previous work \cite{mdm, motiondiffuse, mld}.

\noindent \textbf{$\bullet$ Implementation Details.} On both datasets, we train the MDM and MotionDiffuse models from scratch for $700K$ and $200K$ steps, respectively. To approximate the noisy motion data $\tilde{\mathbf{x}}$ from $\mathbf{x} \in \mathbb{R}^{N\times D}$, we use noisy ranges $[20, 60]$ and $[20, 40]$ for HumanML3D and KIT-ML, respectively.

\noindent \textbf{$\bullet$ Evaluation Metrics.} As suggested by \citeauthor{humanml3d}, the metrics are based on a text feature extractor and a motion feature extractor jointly trained under contrastive loss to produce feature vectors for matched text-motion pairs. R Precision (top 3) measures the accuracy of the top 3 retrieved descriptions for each generated motion, while the Frechet Inception Distance (FID) is calculated using the motion extractor as the evaluator network. Multimodal Distance measures the average Euclidean distance between the motion feature of each generated motion and the text feature of its corresponding description in the test set. Diversity measures the variance of the generated motions across all action categories, while MultiModality measures the diversity of generated motions within each condition.

\noindent \textbf{$\bullet$ Quantitative Result.} Table~\ref{tab:t2m-quantitative} presents quantitative results of our weakly-supervised MotionMix using MDM and MotionDiffuse backbones, in comparison with their original models that are trained with fully annotated and clean motion sequences. To our surprise, in most settings, MotionMix even improves the motion quality (i.e., FID) and multimodal consistency (i.e., R Precision) upon the fully supervised backbones. For example, on HumanML3D and KIT-ML dataset, MDM (\textbf{MotionMix}) commonly reduces FID by over $0.16$ compare to MDM; this comes with the enhancement of both R Precision and Multimodal Distance. We may attribute this to the better generalizability and robustness by involving noisy data in our MotionMix. On the specifical setting of MotionDiffuse (\textbf{MotionMix}) on HumanML3D, though being inferior to the original MotionDiffuse, our MotionMix maintains competitive performance on par with other fully supervised baselines, such as Language2Pose \cite{Language2Pose}, Text2Gestures \cite{Text2Gestures}, \citeauthor{humanml3d} \cite{humanml3d}.

\begin{figure*}[t!]
\centering
\includegraphics[width=0.99\textwidth]{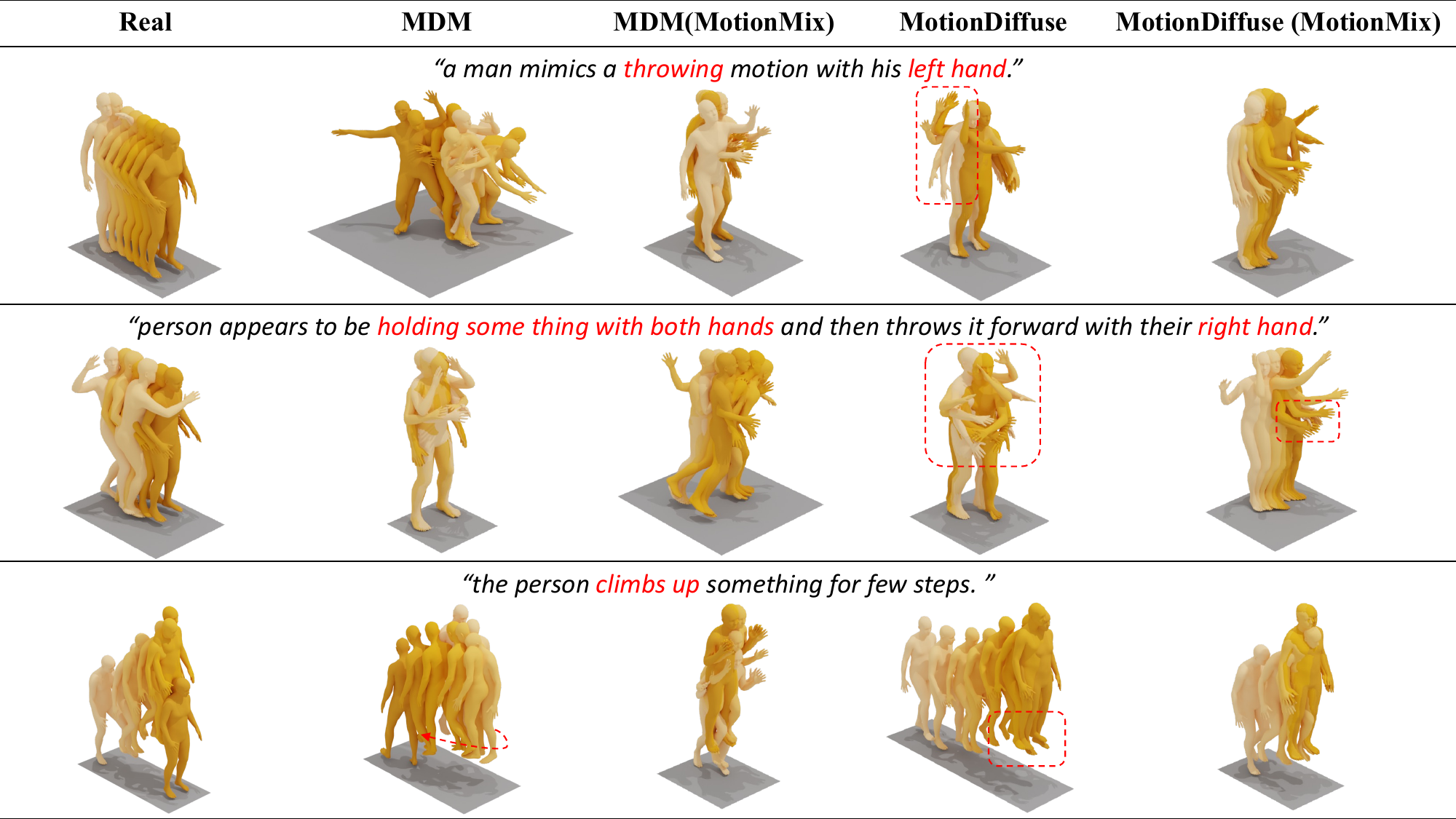}
\caption{
    Qualitative performance of baseline MDM and MotionDiffuse models, trained exclusively on high-quality annotated data, with our MotionMix approach, which learns from imperfect data sources. Their visualized motion results are presented alongside real references for three distinct text prompts. Please refer to supplementary files for more animations.
}
\label{fig:compare-t2m}
\end{figure*}

\begin{table*}
\centering
\small{\resizebox{0.99\textwidth}{!}{
    \begin{tabular}{M{0.01cm}p{3.7cm}M{2.4cm}M{2.5cm}M{2.4cm}M{2.7cm}M{2.5cm}}
    \toprule
    
    & \multirow{2}{3.7cm}{Method} & \multirow{2}{2.0cm}{\centering R Precision (top 3)$\uparrow$} & \multirow{2}{2.5cm}{\centering FID$\downarrow$} & \multirow{2}{2.0cm}{\centering Multimodal Dist.$\downarrow$} & \multirow{2}{2.7cm}{\centering Diversity$\rightarrow$} & \multirow{2}{2.5cm}{\centering Multimodality$\uparrow$} \\ \\
    
    \midrule
    
    \multirow{11}{*}{\rotatebox[origin=c]{90}{HumanML3D}} & Real Motion & $0.797^{\pm.002}$ & $0.002^{\pm.000}$ & $2.974^{\pm.008}$ & $9.503^{\pm.065}$ & - \\
    \cmidrule(lr){2-7}
            & Language2Pose  & $0.486^{\pm.002}$ & $11.02^{\pm.046}$ & $5.296^{\pm.008}$ & $7.676^{\pm.058}$ & - \\
            & Text2Gestures  & $0.345^{\pm.002}$ & $7.664^{\pm.030}$ & $6.030^{\pm.008}$ & $6.409^{\pm.071}$ & - \\
            & \citeauthor{humanml3d} & $0.740^{\pm.003}$ & $1.067^{\pm.002}$ & $3.340^{\pm.008}$ & $9.188^{\pm.002}$ & $2.090^{\pm.083}$ \\
            & MLD  & $0.772^{\pm.002}$ & $0.473^{\pm.013}$ & $3.196^{\pm.010}$ & $9.724^{\pm.082}$ & $2.413^{\pm.079}$ \\
    \cmidrule(lr){2-7}
            & MDM & $0.611^{\pm.007}$ & $0.544^{\pm.440}$ & $5.566^{\pm.027}$ & $9.559^{\pm.860}$ & $2.799^{\pm.072}$ \\
            & MDM (\textbf{MotionMix}) & $0.632^{\pm.006}$ \ua{$3.4\%$} & $0.381^{\pm.042}$ \ua{$30.0\%$} & $5.325^{\pm.026}$ \ua{$4.3\%$} & $9.520^{\pm.090}$ \ua{$69.6\%$} & $2.718^{\pm.019}$ \da{$2.9\%$} \\
    \cmidrule(lr){2-7}
            & MotionDiffuse & $0.782^{\pm.001}$ & $0.630^{\pm.001}$ & $3.113^{\pm.001}$ & $9.410^{\pm.049}$ & $1.553^{\pm.042}$ \\
            & MotionDiffuse (\textbf{MotionMix}) & $0.738^{\pm.006}$ \da{5.6\%} & $1.021^{\pm.071}$ \da{62.1\%} & $3.310^{\pm.020}$ \da{$6.3\%$} & $9.297^{\pm.083}$ \da{$121.5\%$} & $1.523^{\pm.153}$ \da{$1.9\%$} \\

    \midrule
    \midrule

    \multirow{11}{*}{\rotatebox[origin=c]{90}{KIT-ML}} & Real Motion & $0.779^{\pm.006}$ & $0.031^{\pm.004}$ & $2.788^{\pm.012}$ & $11.080^{\pm.097}$ & - \\
    \cmidrule(lr){2-7}
            & Language2Pose & $0.483^{\pm.005}$ & $6.545^{\pm.072}$ & $5.147^{\pm.030}$ & $9.073^{\pm.100}$ & - \\
            & Text2Gestures & $0.338^{\pm.004}$ & $12.12^{\pm.183}$ & $6.964^{\pm.029}$ & $9.334^{\pm.079}$ & - \\
            & \citeauthor{humanml3d} & $0.693^{\pm.007}$ & $2.770^{\pm.109}$ & $3.401^{\pm.008}$ & $10.910^{\pm.119}$ & $1.482^{\pm.065}$ \\
            & MLD  & $0.734^{\pm.007}$ & $0.404^{\pm.027}$ & $3.204^{\pm.027}$ & $10.800^{\pm.117}$ & $2.192^{\pm.071}$ \\
    \cmidrule(lr){2-7}
            & MDM & $0.396^{\pm.004}$ & $0.497^{\pm.021}$ & $9.191^{\pm.022}$ & $10.847^{\pm.109}$ & $1.907^{\pm.214}$ \\
            & MDM (\textbf{MotionMix}) & $0.404^{\pm.005}$ \ua{$2.0\%$} & $0.322^{\pm.020}$ \ua{$35.2\%$} & $9.068^{\pm.019}$ \ua{$1.3\%$} & $10.781^{\pm.098}$ \da{$28.3\%$} & $1.946^{\pm.019}$ \ua{$2.0\%$} \\
    \cmidrule(lr){2-7}
            & MotionDiffuse & $0.739^{\pm.004}$ & $1.954^{\pm.062}$ & $2.958^{\pm.005}$ & $11.100^{\pm.143}$ & $0.730^{\pm.013}$ \\
            & MotionDiffuse (\textbf{MotionMix}) & $0.742^{\pm.005}$ \ua{$0.4\%$} & $1.192^{\pm.073}$ \ua{$39.0\%$} & $3.066^{\pm.018}$ \da{$3.6\%$} & $10.998^{\pm.072}$ \da{$310\%$} & $1.391^{\pm.111}$ \ua{$90.5\%$} \\
            
    \bottomrule
    \end{tabular}
}}
\caption{
    Quantitative results of text-to-motion on the test set of HumanML3D and KIT-ML. Note all baselines are trained with gold data. We run all the evaluation 20 times (except \textit{Multimodality} runs 5 times) and $\pm$ indicates the 95\% confidence interval. $\uparrow$ means higher is better, $\downarrow$ means lower is better, $\rightarrow$ means closer to the real distribution is better. The $\uparrow x\%$ and $\downarrow x\%$ indicate the percentage difference in performance improvement or deterioration when comparing our approach to its correspond baseline.
}
\label{tab:t2m-quantitative}
\end{table*}

\subsection{Action-to-motion}
\label{ssec:a2m}

\noindent \textbf{$\bullet$ Datasets.} We evaluate our MotionMix on two benchmarks: HumanAct12 \cite{humanact} and UESTC \cite{uestc}. HumanAct12 offers 1,191 motion clips categorized into 12 action classes, while UESTC provides 24K sequences of 40 action classes. For this task, we use the pre-processed sequences provided by \citeauthor{actor} as the gold clean motion sequences, and further process them to approximate noisy samples. A pose sequence of $N$ frames is represented in the 24-joint SMPL format \cite{smpl}, using the 6D rotation \cite{Zhou2018OnTC} for every joint, resulting in $\mathbf{p} \in \mathbb{R}^{N\times24\times6}$. A single root translation $\mathbf{r} \in \mathbb{R}^{N\times1\times3}$ is padded and concatenated with $\mathbf{p}$ to obtain the final motion representation $\mathbf{x} = \text{Concat}([\mathbf{p}, \mathbf{r}]) \in \mathbb{R}^{N\times25\times6}$.

\noindent \textbf{$\bullet$ Implementation Details.} Following the experimental setup by \citeauthor{mdm}, we train the MDM (\textbf{MotionMix}) from scratch on the HumanAct12 and UESTC datasets for $750K$ and $2M$ steps, respectively. In our approximation preprocess, we determine the amount of noise to be injected into both the pose sequence $\mathbf{p}$ and the root translation $\mathbf{r}$ by randomly sampling from range $[10, 30]$. The resulting $\tilde{\mathbf{p}}$ and $\tilde{\mathbf{r}}$ are then concatenated to obtain noisy motion $\tilde{\mathbf{\mathbf{x}}}$.

\noindent \textbf{$\bullet$ Evaluation Metrics.} Four metrics are used to assess the quality of generated motions. The FID is commonly used to evaluates the overall quality of generated motions. Accuracy measures the correlation between the generated motion and its action class. Diversity and MultiModality are similar to the text-to-motion metrics.

\noindent \textbf{$\bullet$ Quantitative Result.} Table~\ref{tab:a2m-quantitative} presents the performance outcomes of MDM (\textbf{MotionMix}) and several baseline models, including Action2Motion \cite{humanact}, ACTOR \cite{actor}, INR \cite{Cervantes2022ImplicitNR}, MLD \cite{mld}, and MDM \cite{mdm}, on both the HumanAct12 and UESTC datasets. Following the methodology of \citeauthor{mdm}, we perform 20 evaluations, each comprising 1000 samples, and present average scores with a confidence interval of 95\%. The results highlight that our MotionMix achieves competitive performance with significantly fewer high-quality annotated data instances. In particular, the improvement seen on the UESTC dataset underscores its efficacy in training with noisy motion data from the real-world scenario. On the other hand, the deterioration in performance on HumanAct12 suggests that our approach is better suited for larger datasets, given that the size of HumanAct12 is remarkably smaller than that of UESTC. Nevertheless, our supplementary videos demonstrate that the model trained on HumanAct12 remains capable of generating quality motion sequences based on the provided action classes.

\begin{table}[t!]
\centering
\small{\resizebox{0.99\columnwidth}{!}{
    \begin{tabular}{M{0.03cm}p{2.6cm}M{2.5cm}M{2.4cm}M{2.6cm}M{2.6cm}}
    \toprule
    
    & Method & FID $\downarrow$ & Accuracy $\uparrow$ & Diversity $\rightarrow$ & MultiModality $\rightarrow$ \\
    
    \midrule
    
    \multirow{7}{*}{\rotatebox[origin=c]{90}{HumanAct12}} & Real Motion & $0.053^{\pm.003}$ & $0.995^{\pm.001}$ & $6.835^{\pm.045}$ & $2.604^{\pm.040}$ \\
    \cmidrule(lr){2-6}
            & Action2Motion & $0.338^{\pm.015}$ & $0.917^{\pm.001}$ & $6.850^{\pm.050}$ & $2.511^{\pm.023}$ \\
            & ACTOR & $0.120^{\pm.000}$ & $0.955^{\pm.008}$ & $6.840^{\pm.030}$ & $2.530^{\pm.020}$ \\
            & INR & $0.088^{\pm.004}$ & $0.973^{\pm.001}$ & $6.881^{\pm.048}$ & $2.569^{\pm.040}$ \\
            & MLD & $0.077^{\pm.004}$ & $0.964^{\pm.002}$ & $6.831^{\pm.050}$ & $2.824^{\pm.038}$ \\
    \cmidrule(lr){2-6}
            & MDM & $0.100^{\pm.000}$ & $0.990^{\pm.000}$ & $6.860^{\pm.050}$ & $2.520^{\pm.010}$ \\
            & MDM (\textbf{MotionMix}) & $0.196^{\pm.007}$ \da{$96\%$} & $0.930^{\pm.003}$ \da{$6.1\%$} & $6.836^{\pm.062}$ \ua{$96\%$} & $3.043^{\pm.054}$ \da{$422.6\%$} \\
            
    \midrule
    \midrule
    
    \multirow{6}{*}{\rotatebox[origin=c]{90}{UESTC}} & Real Motion & $2.790^{\pm.290}$ & $0.988^{\pm.001}$ & $33.349^{\pm.320}$ & $14.160^{\pm.060}$ \\
    \cmidrule(lr){2-6}
            % & Action2Motion & - & - & - & - \\
            & ACTOR & $23.430^{\pm2.200}$ & $0.911^{\pm.003}$ & $31.960^{\pm.330}$ & $14.520^{\pm.090}$ \\
            & INR & $15.000^{\pm.090}$ & $0.941^{\pm.001}$ & $31.590^{\pm.190}$ & $14.680^{\pm.070}$ \\
            & MLD & $15.790^{\pm.079}$ & $0.954^{\pm.001}$ & $33.520^{\pm.140}$ & $13.570^{\pm.060}$ \\
    \cmidrule(lr){2-6}
            & MDM & $12.810^{\pm1.460}$ & $0.950^{\pm.000}$ & $33.100^{\pm.290}$ & $14.260^{\pm.120}$ \\
            & MDM (\textbf{MotionMix}) & $11.400^{\pm.393}$ \ua{$11\%$} & $0.960^{\pm.003}$ \ua{$1.1\%$} & $32.806^{\pm.176}$ \da{$118\%$} & $14.277^{\pm.094}$ \da{$17\%$} \\
            
    \bottomrule
    \end{tabular}
}}
\caption{
    Quantitative results of action-to-motion on the HumanAct12 dataset and UESTC test set. We run the evaluation 20 times, and the metric details are similar to Table~\ref{tab:t2m-quantitative}.
}
\label{tab:a2m-quantitative}
\end{table}

\subsection{Music-to-dance}
\label{ssec:m2d}

\noindent \textbf{$\bullet$ Datasets.} We utilize the AIST++ dataset \cite{aistpp}, which comprises 1,408 high-quality dance motions accompanied by music from a diverse range of genres. Following the experimental setup proposed by \citeauthor{edge}, we adopt a configuration in which all training samples are trimmed to 5 seconds and 30 FPS. Similarly to the action-to-motion data, we concatenate $N$-frame pose sequences denoted as $\mathbf{p} \in \mathbb{R}^{N\times24\times6=N\times144}$, along with a single root translation denoted as $\mathbf{r} \in \mathbb{R}^{N\times3}$, and an additional binary contact label for the heel and toe of each foot denoted as $\mathbf{b} \in \{0, 1\}^{N\times4}$. Consequently, EDGE is trained using the final motion representation $\mathbf{x} = \text{Concat}([\mathbf{b}, \mathbf{r}, \mathbf{p}]) \in \mathbb{R}^{N\times151}$.

\noindent \textbf{$\bullet$ Implementation Details.} Similar to the action-to-motion task, we inject noise into both $\mathbf{p}$ and $\mathbf{r}$ using the same noise timestep sampled from $[20, 80]$. Since the contact label $\mathbf{b}$ is obtained from both $\mathbf{p}$ and $\mathbf{r}$, it is not necessary to inject noise into $\mathbf{b}$. Following the setup of \citeauthor{edge}, we train both the EDGE model and our EDGE (\textbf{MotionMix}) from scratch on AIST++ for $2000$ epochs.

\noindent \textbf{$\bullet$ Evaluation Metrics.} To evaluate the quality of the generated dance, we adopt the same evaluation settings as suggested in paper EDGE, including Physical Foot Contact (PFC), Beat Alignment, and Diversity. PFC is a physically-inspired metric that evaluates physical plausibility by capturing realistic foot-ground contact without explicit physical modeling or assuming static contact. Following the previous works \cite{aistpp, bailando}, Beat Alignment evaluates the tendency of generated dances to follow the beat of the music, while Diversity measures the distribution of generated dances in the \quotes{kinetic} ($\text{Dist}_k$) and \quotes{geometric} ($\text{Dist}_g$) feature spaces.

\noindent \textbf{$\bullet$ Quantitative Result.} In contrary to prior works, which typically reported only a single evaluation result, we have observed that the metrics can be inconsistent. Thus, to offer a more comprehensive evaluation, we present the average and 95\% confidence interval, derived from 20 evaluation runs for our retrained EDGE model and our EDGE (\textbf{MotionMix}) variant. For Bailando \cite{bailando} and FACT \cite{aistpp}, we directly fetched results from the paper EDGE \cite{edge}. The results in Table~\ref{tab:m2d-quantitative} vividly demonstrate that, our EDGE (\textbf{MotionMix}) significantly outperforms the baseline across all metrics, showcasing improvements of up to $43.1\%$ in PFC and $95.0\%$ in $\text{Dist}_k$. This further reinforces the generalizability prowess of our MotionMix approach, consistent with the outcomes observed in our text-to-motion experiments.

\begin{table}[t!]
\centering
\small{\resizebox{0.99\columnwidth}{!}{
    \begin{tabular}{p{2.7cm}M{2.5cm}M{2.5cm}M{2.8cm}M{2.5cm}}
    \toprule
    Method & PFC $\downarrow$ & Beat Align. $\uparrow$ & $\text{Dist}_k$ $\rightarrow$ & $\text{Dist}_g$ $\rightarrow$ \\
    \midrule
    Real Motion & 1.380 & 0.314 & 9.545 & 7.766 \\
    Bailando & 1.754 & 0.23 & 10.58 & 7.72 \\
    FACT & 2.2543 & 0.22 & 10.85 & 6.14 \\
    \midrule
    EDGE$\dagger$ & $1.605^{\pm.224}$ & $0.224^{\pm.025}$ & $5.549^{\pm.783}$ & $4.831^{\pm.752}$ \\
    EDGE (\textbf{MotionMix}) & $1.988^{\pm.120}$ \ua{$43.1\%$} & $0.256^{\pm.013}$ \ua{$13.3\%$} & $10.103^{\pm2.039}$ \ua{$95.0\%$} & $6.595^{\pm.173}$ \ua{$15.1\%$} \\   
    \bottomrule
    \end{tabular}
}}
\caption{
    Quantitative results of music-to-dance on the AIST++ test set. We run the evaluation 20 times, and the metric details are similar to Table~\ref{tab:t2m-quantitative}. $\dagger$ denotes the EDGE model that is re-trained by us.\protect\footnotemark
}
\label{tab:m2d-quantitative}
\end{table}

\section{Ablation Studies}
\label{sec:ablation}
MotionMix is introduced as a potential solution that enables the diffusion model to effectively leverage both noisy motion sequences and unannotated data. To demonstrate the efficacy of this approach, we approximate noisy samples from existing datasets and train the model on them, which incorporate several essential hyperparameters: (1) the denoising pivot $T^*$; (2) the ratio of noisy and clean data for training; (3) the noisy range $[T_1, T_2]$ to approximate noisy data. In this section, we thoroughly assess the impact of each hyperparameters within MotionMix. All ablation experiments are carried out on the HumanML3D dataset using the MDM model with the identical settings described in Subsection~\ref{ssec:t2m}.

\footnotetext{The results of the EDGE baseline are different from the ones submitted to AAAI'24 due to a multi-gpu bug. However, our EDGE (MotionMix) still achieves overall better performance.}

\subsection{Effect of The Denoising Pivot $T^*$}
\label{ssec:ablation-denoising-pivot}
We begin our ablation studies by examining the impact of the denoising pivot $T^*$. To evaluate its impact, we conduct experiments with a fixed noisy range of $[T_1, T_2] = [20, 60]$, a noisy ratio of 50\%, and evaluate various $T^*$ values, encompassing $20$, $40$, $60$, and $80$. The results, detailed in Table~\ref{tab:ablation-denoising-pivot}, reveal a notable observation: a roughly estimated denoising pivot is sufficient for real-world scenarios, as evidenced by the competitive outcomes across various $T^*$ values. This robustness underlines the versatility of our MotionMix approach. Additionally, selecting a very small denoising pivot (e.g., $T^* = 0$ or $20$) enables conditions to steer the model toward diverse rough motion sequences before the refining phase, as reflected in the MModality score trend. However, this small value may potentially compromise motion quality, leading to subpar results in other metrics. In contrast, the choice of $T^* = 60$, which is well aligned with our predefined noisy range, yields superior results in multiple evaluation metrics. This sheds light on the need of tuning the denoising pivot to optimize the results, as this hyperparameter determines the starting point for the diffusion model to transform initial noisy motion into high-quality sequences.

\begin{table}[t!]
\centering
\small{\resizebox{0.99\columnwidth}{!}{
    \begin{tabular}{p{3.8cm}M{1.5cm}M{1.3cm}M{1.55cm}M{1.3cm}M{1.85cm}}
    \toprule
    \multirow{2}{3.8cm}{Method} & \multirow{2}{1.5cm}{\centering R Precision (top 3)$\uparrow$} & \multirow{2}{1.2cm}{\centering FID$\downarrow$} & \multirow{2}{1.55cm}{\centering Multimodal Dist.$\downarrow$} & \multirow{2}{1.3cm}{\centering Diversity$\rightarrow$} & \multirow{2}{1.7cm}{\centering Multimodality$\uparrow$} \\ \\
    \midrule
    Real Motion & $0.797^{\pm.002}$ & $0.002^{\pm.000}$ & $2.974^{\pm.008}$ & $9.503^{\pm.065}$ & - \\
    MDM \cite{mdm} & $\underline{0.611^{\pm.007}}$ & $0.544^{\pm.440}$ & $5.566^{\pm.027}$ & $\underline{9.559^{\pm.860}}$ & $2.799^{\pm.072}$ \\
    \midrule
    \textit{50\% noisy, $T_1$=20, $T_2$=60} \\
    MDM (\textbf{MotionMix}) \textit{($T^*$=0)} & $0.598^{\pm.006}$ & $0.714^{\pm.045}$ & $\underline{5.503^{\pm.036}}$ & $9.750^{\pm.123}$ & $\boldsymbol{3.044^{\pm.054}}$ \\
    MDM (\textbf{MotionMix}) \textit{($T^*$=20)} & $0.601^{\pm.005}$ & $0.497^{\pm.048}$ & $5.562^{\pm.026}$ & $9.414^{\pm.092}$ & $\underline{2.935^{\pm.059}}$ \\
    MDM (\textbf{MotionMix}) \textit{($T^*$=40)} & $0.604^{\pm.008}$ & $\underline{0.402^{\pm.032}}$ & $5.524^{\pm.033}$ & $9.396^{\pm.094}$ & $2.747^{\pm.070}$ \\
    MDM (\textbf{MotionMix}) \textit{($T^*$=60)} & $\boldsymbol{0.632^{\pm.006}}$ & $\boldsymbol{0.381^{\pm.042}}$ & $\boldsymbol{5.325^{\pm.026}}$ & $\boldsymbol{9.520^{\pm.090}}$ & $2.718^{\pm.019}$ \\
    MDM (\textbf{MotionMix}) \textit{($T^*$=80)} & $0.594^{\pm.005}$ & $0.589^{\pm.059}$ & $5.670^{\pm.033}$ & $9.242^{\pm.086}$ & $2.602^{\pm.057}$ \\
    \bottomrule
    \end{tabular}
}}
\caption{
    We evaluate MDM (MotionMix) on the HumanML3D test set using different values of the denoising pivot $T^*$. The metrics are calculated in the same manner as detailed in Table~\ref{tab:t2m-quantitative}. The best and the second best result are bold and underlined respectively.
}
\label{tab:ablation-denoising-pivot}
\end{table}

\subsection{Effect of Noisy/Clean Data Ratio}
\label{ssec:ablation-noisy-ratio}
In this ablation study, we evaluate how the noisy/clean data ratio affects our approach by keeping $T^* = 60$ and $[T_1, T_2] = [20, 60]$ constant. We experiment with various noisy ratios of 30\%, 50\%, and 70\%. The results, presented in Table~\ref{tab:ablation-noisy-ratio}, show interesting trends across the evaluation metrics. Notably, higher noisy ratios (i.e., 50\% and 70\%) consistently outperform the lower ratio (i.e., 30\%). Note that, a higher noisy ratio allows the model to access a wider range of annotated text conditions, yielding better R Precision and Multimodal Distance. On the other hand, the 30\% ratio, despite being trained with a greater amount of clean data, exhibits suboptimal motion quality (scoring $0.898$ in FID) in comparison to other supervised baselines in Table~\ref{tab:t2m-quantitative}, such as Language2Pose (FID of $11.02$), Text2Gestures (FID of $7.664$), \citeauthor{humanml3d} (FID of $1.067$). Nevertheless, it still achieves results on par with the supervised MDM baseline in terms of multimodal consistency (i.e. Multimodal Distance). These observations underscore the resilience of our MotionMix approach to variations in the noisy/clean data ratio.

\begin{table}[t!]
\centering
\small{\resizebox{0.99\columnwidth}{!}{
    \begin{tabular}{p{4.35cm}M{1.5cm}M{1.3cm}M{1.55cm}M{1.4cm}M{1.85cm}}
    \toprule
    \multirow{2}{4.35cm}{Method} & \multirow{2}{1.5cm}{\centering R Precision (top 3)$\uparrow$} & \multirow{2}{1.2cm}{\centering FID$\downarrow$} & \multirow{2}{1.55cm}{\centering Multimodal Dist.$\downarrow$} & \multirow{2}{1.4cm}{\centering Diversity$\rightarrow$} & \multirow{2}{1.7cm}{\centering Multimodality$\uparrow$} \\ \\
    \midrule
    Real Motion & $0.797^{\pm.002}$ & $0.002^{\pm.000}$ & $2.974^{\pm.008}$ & $9.503^{\pm.065}$ & - \\
    MDM & $0.611^{\pm.007}$ & $0.544^{\pm.440}$ & $5.566^{\pm.027}$ & $9.559^{\pm.860}$ & $2.799^{\pm.072}$ \\
    \midrule
    \textit{$T_1$=20, $T_2$=60, $T^*$=60} \\
    MDM (\textbf{MotionMix}) \textit{(30\% noisy)} & $0.601^{\pm.007}$ & $0.898^{\pm.045}$ & $5.581^{\pm.030}$ & $9.080^{\pm.092}$ & $\underline{2.856^{\pm.074}}$ \\
    MDM (\textbf{MotionMix}) \textit{(50\% noisy)} & $\boldsymbol{0.632^{\pm.006}}$ & $\underline{0.381^{\pm.042}}$ & $\boldsymbol{5.325^{\pm.026}}$ & $\boldsymbol{9.520^{\pm.090}}$ & $2.718^{\pm.019}$ \\
    MDM (\textbf{MotionMix}) \textit{(70\% noisy)} & $\underline{0.615^{\pm.006}}$ & $\boldsymbol{0.359^{\pm.030}}$ & $\underline{5.545^{\pm.031}}$ & $\underline{9.457^{\pm.098}}$ & $\boldsymbol{2.867^{\pm.107}}$ \\
    \bottomrule
    \end{tabular}
}}
\caption{
    We evaluate MDM (MotionMix) on the HumanML3D test set using different ratios for noisy and clean data. The metrics are calculated in the same manner as detailed in Table~\ref{tab:t2m-quantitative}. The best and the second best result are bold and underlined respectively.
}
\label{tab:ablation-noisy-ratio}
\end{table}

\subsection{Effect of The Noisy Range}
\label{ssec:ablation-noisy-range}
The purpose of the noisy range in our work is to approximate the noise schedule found in real-world motion capture data. Thus, for different datasets in Section~\ref{sec:experiments}, we choose noisy ranges based on the visualization of motion from each dataset. For example, UESTC \cite{uestc} contains noisy mocap data, while HumanML3D \cite{humanml3d}, derived from AMASS \cite{amass}, consists of clean motion sequences. This ablation, therefore, comprehensively evaluates the effectiveness of our MotionMix approach when handling different noisy levels of motion sequences. We categorize the evaluations into two groups: narrow/wide ranges of noise and low/high schedules of noise. All experiments are conducted with a noisy ratio of 50\%, and the denoising pivot $T^*$ is equal to the chosen $T_2$. The results are presented in Table~\ref{tab:ablation-noisy-range}.

\noindent \textbf{$\bullet$ Narrow/Wide Noisy Range.} Three noisy ranges $[T_1, T_2] \in \{[20, 40], [20, 60], [20, 80]\}$ are set to analyze the effect of \textit{how much the range spans}. Counterintuitively, the smaller noisy range does not equal to the better performance. For example, noisy ranging from 20 to 60 time steps leads to overall the best performance, compared to range $[20, 40]$. Though, large noisy range (i.e., $[20, 80]$) unevitably deteriotate the model capacity.

\noindent \textbf{$\bullet$ Low/High Noisy Schedule.} Four contrast ranges $[T_1, T_2] \in \{[10, 30], [20, 40], [40, 60], [60, 80]\}$ are experimented to evaluate the robustness of MotionMix regarding \textit{corruption level} of noisy motions. Notably, our proposed MotionMix performs reasonably stable on different levels of corrupted motions. More visual animations are also provided in our supplementary videos.

\begin{table}[t!]
\centering
\small{\resizebox{0.99\columnwidth}{!}{
    \begin{tabular}{p{4.8cm}M{1.5cm}M{1.3cm}M{1.55cm}M{1.3cm}M{1.85cm}}
    \toprule
    \multirow{2}{4.8cm}{Method} & \multirow{2}{1.5cm}{\centering R Precision (top 3)$\uparrow$} & \multirow{2}{1.2cm}{\centering FID$\downarrow$} & \multirow{2}{1.55cm}{\centering Multimodal Dist.$\downarrow$} & \multirow{2}{1.3cm}{\centering Diversity$\rightarrow$} & \multirow{2}{1.7cm}{\centering Multimodality$\uparrow$} \\ \\
    \midrule
    Real Motion & $0.797^{\pm.002}$ & $0.002^{\pm.000}$ & $2.974^{\pm.008}$ & $9.503^{\pm.065}$ & - \\
    MDM & $0.611^{\pm.007}$ & $0.544^{\pm.440}$ & $5.566^{\pm.027}$ & $9.559^{\pm.860}$ & $2.799^{\pm.072}$ \\
    \midrule
    \textit{50\% noisy, $T^* = T_2$ } \\
    MDM (\textbf{MotionMix}) \textit{($T_1$=20, $T_2$=40)} & $\underline{0.616^{\pm.006}}$ & $\underline{0.451^{\pm.033}}$ & $\underline{5.459^{\pm.027}}$ & $9.585^{\pm.101}$ & $2.585^{\pm.076}$ \\
    MDM (\textbf{MotionMix}) \textit{($T_1$=20, $T_2$=60)} & $\boldsymbol{0.632^{\pm.006}}$ & $\boldsymbol{0.381^{\pm.042}}$ & $\boldsymbol{5.325^{\pm.026}}$ & $\boldsymbol{9.520^{\pm.090}}$ & $\underline{2.718^{\pm.019}}$ \\
    MDM (\textbf{MotionMix}) \textit{($T_1$=20, $T_2$=80)} & $0.604^{\pm.004}$ & $0.614^{\pm.060}$ & $5.540^{\pm.024}$ & $\underline{9.554^{\pm.104}}$ & $\boldsymbol{2.768^{\pm.095}}$ \\
    \midrule
    \textit{50\% noisy, $T^* = T_2$ } \\
    MDM (\textbf{MotionMix}) \textit{($T_1$=10, $T_2$=30)} & $0.592^{\pm.008}$ & $0.713^{\pm.048}$ & $5.633^{\pm.028}$ & $9.567^{\pm.109}$ & $2.783^{\pm.139}$ \\
    MDM (\textbf{MotionMix}) \textit{($T_1$=20, $T_2$=40)} & $\boldsymbol{0.616^{\pm.006}}$ & $\underline{0.451^{\pm.033}}$ & $\boldsymbol{5.459^{\pm.027}}$ & $9.585^{\pm.101}$ & $2.585^{\pm.076}$ \\
    MDM (\textbf{MotionMix}) \textit{($T_1$=40, $T_2$=60)} & $\underline{0.598^{\pm.004}}$ & $0.554^{\pm.076}$ & $5.600^{\pm.031}$ & $\boldsymbol{9.479^{\pm.100}}$ & $\underline{2.815^{\pm.094}}$ \\
    MDM (\textbf{MotionMix}) \textit{($T_1$=60, $T_2$=80)} & $0.597^{\pm.008}$ & $\boldsymbol{0.437^{\pm.039}}$ & $\underline{5.554^{\pm.033}}$ & $\underline{9.452^{\pm.092}}$ & $\boldsymbol{2.895^{\pm.079}}$ \\
    \bottomrule
    \end{tabular}
}}
\caption{
    We evaluate MDM (MotionMix) on the HumanML3D test set using different noisy ranges $[T_1, T_2]$ to approximate the noisy motion sequences. The table presents two distinct scenarios: the upper block ablates \textit{how much the range spans}, while the lower block examines the impact of the \textit{corruption level} of noisy motions. The metrics are calculated in the same manner as detailed in Table~\ref{tab:t2m-quantitative}. For each setting, the best and the second best result are bold and underlined respectively.
}
\label{tab:ablation-noisy-range}
\end{table}

\section{Conclusion}
\label{sec:conclusion}
In this work, we look into the realm of conditional human motion generation, devling into the challenge of training with both \textit{noisy annotated} and \textit{clean unannotated} motion sequences. The proposed approach, \textbf{MotionMix}, pioneers the utilization of a weakly-supervised diffusion model as a potential solution for this challenge. This innovative method effectively overcomes the constraints arising from limited high-quality annotated data, achieving competitive results compared to fully supervised models. The versatility of MotionMix is showcased across multiple motion generation benchmarks and fundamental diffusion model designs. Comprehensive ablation studies further bolster its resilience in diverse noisy schedules and the strategic selection of the denoising pivot.

\appendix

\begin{table*}[ht!]
\centering
\resizebox{0.9\textwidth}{!}{
    \begin{tabular}{p{4.6cm}M{2.0cm}M{2.0cm}M{2.0cm}M{2.0cm}}
    \toprule
    Method & PFC $\downarrow$ & Beat Align. $\uparrow$ & $\text{Dist}_k$ $\rightarrow$ & $\text{Dist}_g$ $\rightarrow$ \\
    \midrule
    Real Motion (AIST++) & 1.380 & 0.314 & 9.545 & 7.766 \\
    Real Motion (AMASS) & 1.032 & - & - & - \\
    EDGE$\dagger$ & $1.605^{\pm.224}$ & $0.224^{\pm.025}$ & $\underline{5.549^{\pm.783}}$ & $\underline{4.831^{\pm.752}}$ \\
    
    \midrule
    
    \multicolumn{5}{l}{\textit{Half noisy AIST++ and half clean AIST++ (in our main paper)}} \\
    EDGE (\textbf{MotionMix})  & $1.988^{\pm.120}$ & $\boldsymbol{0.256^{\pm.013}}$  & $\boldsymbol{10.103^{\pm2.039}}$ & $\boldsymbol{6.595^{\pm.173}}$ \\

    \midrule
    
    \multicolumn{5}{l}{\textit{Combine clean AIST++ and clean AMASS}} \\
    EDGE (\textbf{MotionMix}) \textit{($T^*$=20)} & $\underline{1.310^{\pm.078}}$ & $0.236^{\pm.007}$ & $3.437^{\pm.229}$ & $4.308^{\pm.134}$ \\    
    EDGE (\textbf{MotionMix}) \textit{($T^*$=40)} & $\boldsymbol{1.062^{\pm.080}}$ & $\underline{0.240^{\pm.009}}$ & $3.639^{\pm.292}$ & $4.371^{\pm.111}$ \\
    \bottomrule
    \end{tabular}
}
\caption{
    Quantitative results of music-to-dance on the AIST++ test set. We run the evaluation 20 times. The best and the second best result are bold and underlined respectively. $\dagger$ denotes the EDGE model that is re-trained by us. 
}

\label{tab:real-case-m2d-tmp}
\end{table*}

\section{Application - Real Case Scenario}
We experimented training the EDGE model using both AIST++ and AMASS together. With AMASS (low PFC), our model can generate plausible motion with less skating (PFC: 1.06, Tab.~\ref{tab:real-case-m2d-tmp}), visually supported by videos on our project page.

\section{Application - Motion Editing}
MDM \cite{mdm} introduced two motion editing applications: \textbf{in-betweening} and \textbf{body part editing}. These applications share the same approach, respectively, in the temporal and spatial domains. For \textbf{in-betweening}, they maintained the initial and final 25\% of the motion sequence as fixed, while the model generated the intermediate 50\%. In the context of \textbf{body part editing}, specific joints were held fixed, leaving the model responsible for generating the remaining segments. In particular, their experimentation focused on editing the upper body joints exclusively. In our supplementary videos, we demonstrate that, in both scenarios, our MDM (\textbf{MotionMix}) does not compromise this useful feature, exhibiting the ability to produce coherence motion sequences that align with both the motion's fixed section and the given condition (if provided).

\bibliography{aaai24}

\end{document}